\documentclass[runningheads]{llncs}

\usepackage[mobile]{eccv}

\usepackage{eccvabbrv}

\usepackage{graphicx}
\usepackage{booktabs}

\usepackage[accsupp]{axessibility}  % Improves PDF readability for those with disabilities.

\newcommand{\mypar}[1]{\par\addvspace{0.4em}\noindent\textbf{#1}}

\newcommand{\improve}[1]{ \raisebox{0.4ex}{ \tiny \textcolor{teal}{($+$#1)} } }
\newcommand{\worse}[1]{ \raisebox{0.4ex}{\tiny \textcolor{gray}{($-$#1)}} }
\newcommand{\OURS}{Diff3R}

\usepackage{hyperref}

\usepackage{orcidlink}

\begin{document}

% ---------------------------------------------------------------
\title{\OURS: Feed-forward 3D Gaussian Splatting with Uncertainty-aware Differentiable Optimization} 

\titlerunning{Diff3R}
\authorrunning{Y.-C. Liu et al.}

% Springer Heidelberg, Tiergartenstr.~17, 69121 Heidelberg, Germany

\author{Yueh-Cheng Liu\inst{1} \and
Jozef Hladk\'{y}\inst{2} \and
Matthias Nie{\ss}ner\inst{1} \and
Angela Dai\inst{1}}

\institute{Technical University of Munich \and
Computing Systems Lab,  Huawei Technologies, Switzerland \\
\url{https://liu115.github.io/diff3r}
}
\maketitle

% \begin{abstract}
% Feed-forward 3DGS has shown huge improvement recently.
% However, 
%   \keywords{First keyword \and Second keyword \and Third keyword}
% \end{abstract}

\begin{abstract}
Recent advances in 3D Gaussian Splatting (3DGS) present two main directions: feed-forward models offer fast inference in sparse-view settings, while per-scene optimization yields high-quality renderings but is computationally expensive. 
To combine the benefits of both, we introduce \OURS{}, a novel framework that explicitly bridges feed-forward prediction and test-time optimization. By incorporating a differentiable 3DGS optimization layer directly into the training loop, our network learns to predict an optimal initialization for test-time optimization rather than a conventional zero-shot result.
To overcome the computational cost of backpropagating through the optimization steps, we propose computing analytical gradients via the Implicit Function Theorem and a scalable, matrix-free PCG solver tailored for 3DGS optimization.
Additionally, we incorporate a data-driven uncertainty model into the optimization process by adaptively controlling how much the parameters are allowed to change during optimization. This approach effectively mitigates overfitting in under-constrained regions and increases robustness against input outliers. 
Since our proposed optimization layer is model-agnostic, we show that it can be seamlessly integrated into existing feed-forward 3DGS architectures for both pose-given and pose-free methods, providing improvements for test-time optimization.
\end{abstract}

\section{Introduction}
\label{sec:intro}

Novel view synthesis (NVS) from a set of unposed or posed images is a fundamental and longstanding challenge in 3D computer vision. Recently, 3D Gaussian Splatting (3DGS)~\cite{kerbl20233d} has revolutionized the field by achieving real-time, high-fidelity rendering through an unstructured, explicit point-based representation. However, the standard 3DGS pipeline requires computationally expensive, per-scene optimization, limiting its applicability. To circumvent this, a recent wave of feed-forward 3DGS frameworks has emerged. By leveraging large-scale datasets, these models learn to directly predict 3D Gaussian parameters from multiple input views, operating in both pose-given~\cite{charatan2024pixelsplat, chen2024mvsplat, liu2024mvsgaussian, zhang2024gs, xu2025depthsplat, kang2025ilrm} and pose-free~\cite{ye2025yonosplat, ye2024no, lin2025depth, jiang2025anysplat, smart2024splatt3r} settings, even with only a few images.

While feed-forward models demonstrate impressive generalization and fast inference, they fundamentally rely on learned priors. Consequently, they often struggle to reconstruct scene-specific, high-frequency details, leading to blurry or over-smoothed renderings. A natural strategy to bridge this domain gap is Test-Time Optimization (TTO), where the predicted Gaussians are fine-tuned using the available context views. However, in the highly under-constrained sparse-view setting, standard TTO frequently falls into a critical failure mode: catastrophic overfitting. Because the optimization lacks sufficient multi-view supervision, the model can easily manipulate the Gaussians to perfectly reconstruct the context views while severely corrupting the underlying geometry (\eg, generating floaters or cheated view-dependent effects), drastically degrading novel view synthesis performance.

To address this issue, we propose \OURS{}, a novel method to train feed-forward 3DGS models. Instead of training a network to predict the final set of 3D Gaussians directly, we propose training it explicitly to predict the optimal initialization for test-time optimization. Drawing inspiration from meta-learning and hyperparameter optimization~\cite{finn2017model, franceschi2018bilevel}, we formulate this as a bilevel optimization problem. During training, our model predicts an initial set of Gaussians, which undergo a brief inner-loop optimization to fit the context views. The network weights are then updated in the outer loop based on the novel-view rendering quality of the optimized Gaussians.

Key to our success is efficiently backpropagating through this optimization during training. To solve this, we propose a tailored, differentiable optimization strategy. By leveraging the Implicit Function Theorem and Gauss-Newton approximations, we analytically compute the implicit gradients~\cite{rajeswaran2019meta, lorraine2020optimizing} of the outer novel-view loss with respect to the initial Gaussian parameters. This formulation completely avoids unrolling the optimization trajectory (\eg, differentiating through every optimization step), enabling memory-efficient, end-to-end training of the entire pipeline.

Furthermore, to explicitly tackle the overfitting problem during TTO, we introduce an uncertainty-aware optimization mechanism. 
% We observe that feed-forward predictions are not uniformly reliable, as some Gaussians suffer from uncertainty (\eg, due to network limitations). 
We observe that feed-forward predictions are not uniformly reliable for downstream optimization.
Therefore, \OURS{} is designed to predict learnable per-Gaussian, per-attribute regularization weights (\ie, uncertainty), alongside the initial 3D Gaussians. These weights dynamically regulate the inner optimization: the model learns to penalize deviations from the feed-forward prior strictly for highly confident predictions, while granting heavily uncertain Gaussians the freedom to adapt. Remarkably, because our implicit gradients flow back through these meta-parameters, the network learns to quantify its own uncertainty purely from outer-loop novel-view supervision, requiring no ground-truth uncertainty labels.

Since our proposed method is model agnostic, we build our model based on DepthSplat~\cite{xu2025depthsplat} and Depth Anything v3 \cite{lin2025depth} to support both pose-given and pose-free settings. Our method achieves improvements over feed-forward baselines after test-time optimization, gaining over 0.25 dB PSNR in RealEstate10K and 0.35 dB PSNR on ScanNet++. Ablation studies confirm that both the differentiable optimization framework and the learned uncertainty regularization are essential for these gains.

In summary, our main contributions are as follows:
\begin{itemize}
    \item We propose a novel bilevel optimization framework for feed-forward 3DGS. By analytically computing implicit gradients via a matrix-free approach, we enable efficient, end-to-end backpropagation through the test-time optimization process.
    \item We introduce a adaptive uncertainty modeling strategy. By predicting per-parameter regularization weights, our model learns to penalize overfitting to sparse context views during optimization, balancing the feed-forward prior with new information from test-time observations.
\end{itemize}

\begin{figure}[tb]
  \centering
  \includegraphics[width=\linewidth]{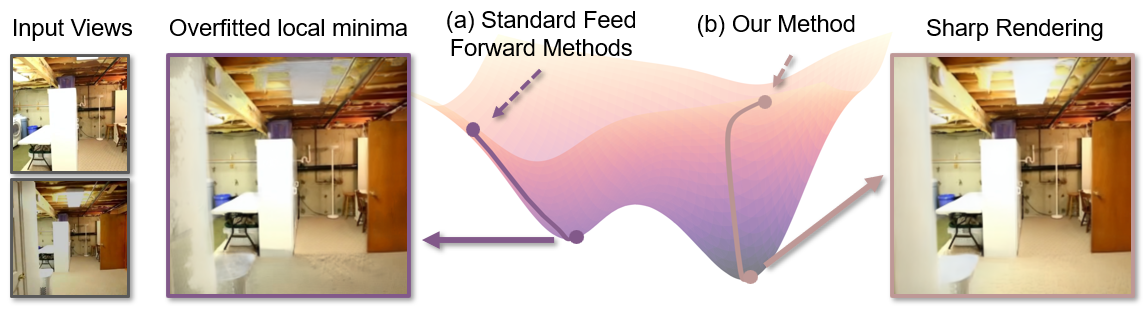}
  \vspace{-5mm}
  \caption{\textbf{Optimization-Aware 3D Gaussian Splatting.} (a) Standard feed-forward 3DGS models directly predict Gaussian parameters, frequently resulting in blurry novel views. Applying test-time optimization (TTO) to these predictions in sparse-view settings is highly unconstrained, which makes it easily overfits to the input views, trapping the solution in poor local minima with severe visual artifacts. (b) By training through the optimization process via implicit gradients, our method learns an optimization-aware initialization, tailored for 3DGS post-optimization. 
  % Concurrently, our model predicts a learnable uncertainty regularizer that bounds the optimization (shaded region), preventing deviation from the learned prior and reliably guiding the TTO to a high-fidelity global minimum, reducing overfitting.
}
  \label{fig:teaser}
\end{figure}

\section{Related Work}

\label{sec:related}

\subsection{Feed-forward 3DGS}
While the original 3D Gaussian Splatting (3DGS)~\cite{kerbl20233d} achieves impressive real-time rendering, it relies on dense multi-view images and time-consuming per-scene optimization. To overcome this, recent works use feed-forward neural networks to predict 3DGS parameters directly from just a few input images. These feed-forward methods can be broadly categorized into two groups based on their reliance on camera information.

\textbf{Pose-given methods}~\cite{charatan2024pixelsplat, chen2024mvsplat, liu2024mvsgaussian, zhang2024gs, xu2025depthsplat, kang2025ilrm} leverage known camera intrinsics and extrinsics to construct multi-view geometry priors. By utilizing epipolar attention mechanisms or 3D cost volumes, these networks aggregate features across views to predict the geometry (means, scales, rotations) and appearance (opacity, spherical harmonics) of the Gaussians for each pixel (\ie, per-pixel Gaussians).

\textbf{Pose-free methods} rely on recent advances in two-view~\cite{wang2024dust3r, leroy2024grounding} or multi-view~\cite{wang2025vggt, keetha2025mapanything, lin2025depth} feed-forward reconstruction models, which directly predict 3D geometry via global attention transformers. Building on these foundations, several approaches~\cite{ye2024no, huang2025no, jiang2025anysplat, ye2025yonosplat, lin2025depth} are now capable of estimating both the camera poses and the Gaussian attributes directly from unposed images.

% However, all these methods are trained to minimize zero-shot reconstruction error, producing predictions that represent suboptimal local minima. When subjected to standard test-time optimization, they suffer catastrophic overfitting in sparse-view settings. We address this by training through the optimization process itself, learning both initialization and uncertainty-aware regularization that guides refinement toward global minima.

However, the methods above are trained to minimize zero-shot reconstruction error. As a result, their predictions often act as poor starting points for further refinement. When subjected to standard test-time optimization, they easily suffer from catastrophic overfitting in sparse-view settings, getting trapped in suboptimal local minima. We address this by training through the optimization process itself and learning an optimization-aware initialization for 3DGS.

% Rather than per-scene optimization, feed-forward methods predict 3D Gaussian primitives directly from sparse views. \textbf{Pose-given} methods~\cite{charatan2024pixelsplat,chen2024mvsplat,xu2025depthsplat,zhang2024gs} assume known camera parameters and leverage cost volumes or transformers for geometry estimation. \textbf{Pose-free} approaches~\cite{ye2024no,jiang2025anysplat,ye2025yonosplat} jointly estimate poses and scene structure via canonical spaces or transformer-based aggregation. However, all these methods are trained to minimize zero-shot reconstruction error, producing predictions that represent suboptimal local minima. When subjected to standard test-time optimization, they suffer catastrophic overfitting in sparse-view settings. We address this by training through the optimization process itself, learning both initialization and uncertainty-aware regularization that guides refinement toward global minima.

\subsection{Differentiable Optimization}

Differentiable optimization enables end-to-end learning by backpropagating through geometric solvers. Prior works have successfully differentiated through Gauss-Newton steps for non-rigid tracking~\cite{liLearningOptimizeNonRigid2020}, Levenberg-Marquardt alignment~\cite{vonstumbergLMRelocLevenbergMarquardtBased2020}, and bundle adjustment~\cite{tangBANetDenseBundle2019, roessle2023e2emultiviewmatching} using implicit differentiation. More recently, VGGSfM~\cite{wang2024vggsfm} extended these concepts to full Structure-from-Motion (SfM) pipelines using differentiable Cholesky solvers \cite{pineda2022theseus}. 

While these methods successfully integrate geometric computation into deep learning for sparse point clouds and camera poses, differentiating through the optimization of 3D Gaussian Splatting remains a largely under-explored area. This is primarily due to the sheer scale of the representation. A typical 3DGS scene consists of hundreds of thousands to millions of Gaussians, each with multiple attributes (position, scale, rotation, opacity, and spherical harmonics). This massive number of parameters, combined with millions of per-pixel rendering residuals, makes standard backpropagation through auto-differentiation expensive and memory-intensive. In this work, we address this gap by deriving a memory-efficient implicit gradient formulation specifically tailored to handle the massive parameter space of 3DGS.

\subsection{Meta-Learning for 3D Reconstruction}

Meta-learning, often framed as ``learning to learn,'' focuses on training models that can rapidly adapt to new tasks using minimal data. A foundational approach in this domain is MAML~\cite{finn2017model}, which employs bilevel optimization to discover an optimal network initialization for fast downstream adaptation. To reduce the memory burden of backpropagating through unrolled gradient steps, subsequent works introduced efficient alternatives by first-order approximation \cite{nichol2018first,nichol2018reptile} or Implicit Function Theorem \cite{rajeswaran2019meta}. Furthermore, methods like Meta-SGD~\cite{li2017meta} expanded this paradigm by learning both the initialization and per-parameter adaptation rates.

% Extensions to neural representations include MetaSDF\cite{sitzmann2019metasdf} for signed distance functions and meta-learned initial weights for NeRFs\cite{tancik2020meta}. Recent works apply similar ideas to 3DGS: G3R\cite{chen2024g3r} and QuickSplat\cite{liu2025quicksplat} learn iterative refinement networks but backpropagate through unrolled optimization steps, requiring memory proportional to the number of iterations. In contrast, we predict optimal initializations for standard test-time optimization and employ implicit differentiation to compute exact gradients at convergence without unrolling, enabling memory-efficient training of uncertainty-aware components.

In the context of 3D vision, these bilevel optimization strategies have been successfully applied to coordinate-based neural representations. For example, MetaSDF~\cite{sitzmann2019metasdf} and meta-learned NeRFs~\cite{tancik2020meta} optimize global initial network weights to speed up per-scene convergence. More recently, similar ideas have been adapted for 3D Gaussian Splatting. Works such as G3R~\cite{chen2024g3r} and QuickSplat~\cite{liu2025quicksplat} train iterative refinement networks to update Gaussian parameters. However, these methods rely on backpropagating directly through unrolled optimization steps or custom networks to predict the updates resulting in a large memory footprint. In contrast, our framework predicts an optimization-aware initialization and utilizes implicit differentiation to compute exact analytical gradients at convergence. This completely bypasses the need to unroll iterations, enabling the memory-efficient, end-to-end training of our adaptive uncertainty regularizers.

% Meta-learning, or ``learning to learn,'' trains models to adapt rapidly to new tasks using minimal data. MAML\cite{finn2017model} learns network initializations for fast adaptation via bilevel optimization, while Implicit MAML\cite{rajeswaran2019meta} uses the Implicit Function Theorem to compute exact meta-gradients at convergence without storing optimization trajectories.

%\input{sec/related_long}

% 
\section{Method}
\label{sec:method}

\begin{figure}[tb]
  \centering
  \includegraphics[width=\linewidth]{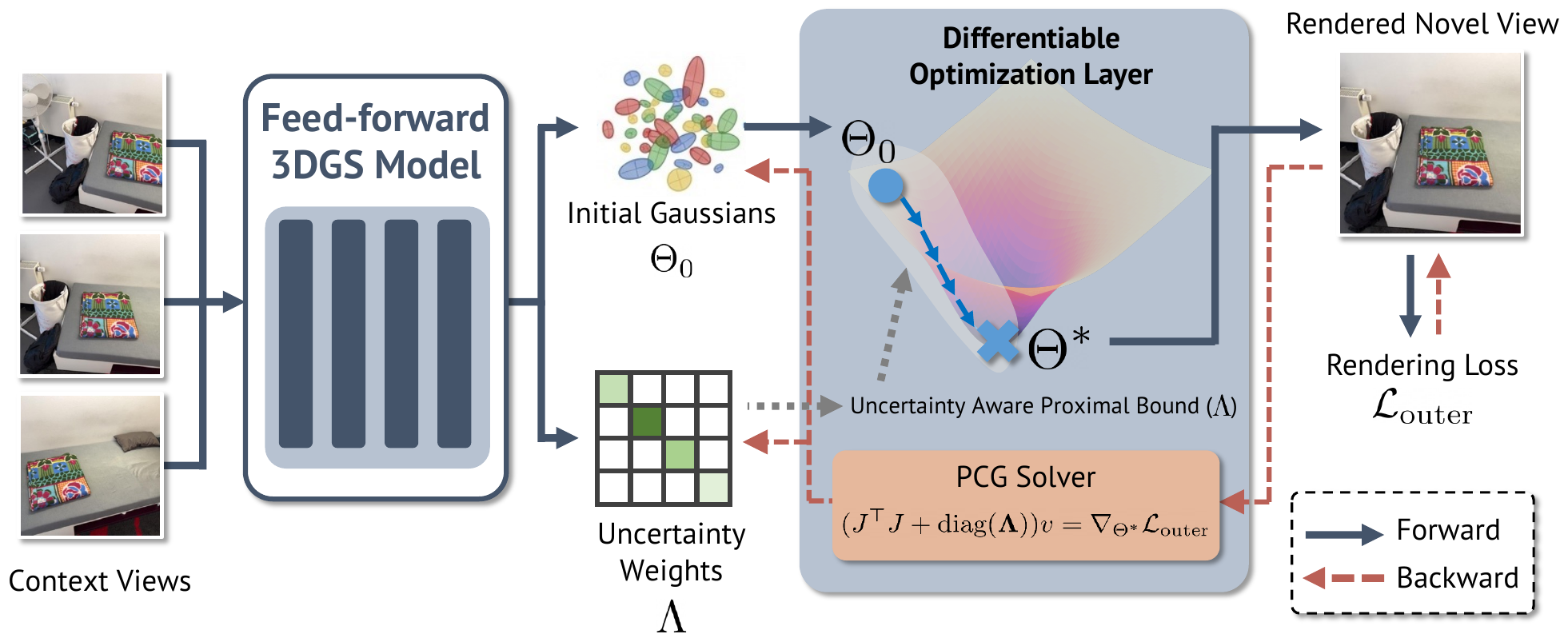}
  \vspace{-5mm}
  \caption{\textbf{Overview of our Uncertainty-Aware Differentiable 3DGS Framework.}  Given a sparse set of context views (with optional camera parameters), our feed-forward network predicts an initial set of 3D Gaussian parameters ($\Theta_0$). Our proposed differentiable optimization layer refines these parameters via gradient descent to yield the optimized Gaussians ($\Theta^*$). To train the network end-to-end, we introduce an efficient analytical solution for the backward pass using implicit gradients and a matrix-free PCG solver. Additionally, to make the optimization more robust in sparse-view settings, we predict learnable uncertainty weights ($\boldsymbol{\Lambda}$). These weights act as an adaptive proximal bound on the optimization trajectory, preventing the model from overfitting to the context views.
}
  \label{fig:method}
\end{figure}

\subsection{Preliminaries: 3D Gaussian Splatting}

\mypar{3D Gaussian splatting}
3D Gaussian Splatting (3DGS) \cite{kerbl20233d} represents a 3D scene using an unstructured set of 3D Gaussians. Each Gaussian is parameterized by a center position (mean) $\mu \in \mathbb{R}^3$, a 3D covariance matrix $\Sigma$, an opacity value $\alpha \in [0, 1]$, and the color $c$, represented by spherical harmonics (SH) coefficients. To ensure positive semi-definiteness during optimization, the covariance matrix is decomposed into a scaling matrix $S$ and a rotation matrix $R$, such that $\Sigma = R S S^\top R^\top$. We denote the full set of 3D Gaussian parameters as $\Theta = \{ \mu_i, S_i, R_i, \alpha_i, c_i \}_{i=1}^N$. 

To render an image from a specific camera pose $P$, the 3D Gaussians are projected onto the 2D image plane. The final pixel color $C$ is computed via volumetric alpha-blending of $N$ sorted overlapping Gaussians:
\begin{equation}
    C = \sum_{i=1}^{N} c_i \alpha'_i \prod_{j=1}^{i-1} (1 - \alpha'_j),
\end{equation}
where $\alpha'_i$ evaluates the 2D projected opacity of the $i$-th Gaussian. For brevity, we denote the entire differentiable rendering process mapping the Gaussian parameters to an image under a camera pose $P$ as $I = f(\Theta, P)$, or simply $f(\Theta)$ when the pose is implicit.

\mypar{Feed-forward 3D Gaussian Splatting.}
Standard 3DGS requires computationally expensive per-scene optimization. Feed-forward 3DGS aims to circumvent this by predicting the parameters $\Theta$ directly from a set of context views $\{I_{ctx}, P_{ctx}\}$. Generally, a neural network $g_\phi$, parameterized by weights $\phi$, extracts image features to build a 3D cost volume or predict per-view depth maps. These are back-projected into 3D space to establish the initial Gaussian means $\mu$. Subsequent network layers predict the remaining attributes (scale, rotation, opacity, and SH coefficients) for each point. We formulate this feed-forward prediction as the initialization for our Gaussians: $\Theta_0 = g_\phi(\{I_{ctx}, P_{ctx}\})$.

\subsection{Differentiable 3DGS Optimization}
While feed-forward models produce fast initializations $\Theta_0$, they often lack high-frequency detail. Our goal is to train a feed-forward 3DGS model that outputs 3D Gaussians designed for downstream test-time optimization. Therefore, we propose a bilevel optimization framework. The \textit{inner loop} performs a brief test-time optimization (TTO) to fit the context views, while the \textit{outer loop} updates the network weights $\phi$ by rendering novel views.

The inner optimization objective minimizes the photometric loss against the context views, regularized by a proximal term. Letting $r(\Theta) = f(\Theta) - I_{ctx}$ denote the rendering residual (\ie, the pixel difference between the rendering and the GT for each pixel), the inner loss is formulated as:
\begin{equation} \label{eq:inner}
    \mathcal{L}_\text{inner}(\Theta) =  \mathcal{L}_\text{photo}(\Theta) +  \mathcal{L}_\text{proximal}(\Theta)  = \frac{1}{2} \| r(\Theta) \|_2^2 + \frac{\lambda}{2} \| \Theta - \Theta_0 \|_2^2,
\end{equation}
where $\lambda$ is a global regularization scalar. We denote the converged parameters of this inner optimization as $\Theta^* = \arg\min_\Theta \mathcal{L}_\text{inner}(\Theta)$.

We include the proximal term $\mathcal{L}_\text{proximal}(\Theta)$ \cite{rajeswaran2019meta} for two critical reasons. First, it bounds the optimization, ensuring the Gaussians do not deviate excessively from the prior learned by the feed-forward network, which is particularly helpful for training differentiable optimization. Secondly, it ensures the optimality condition such that it has gradients w.r.t. the initial $\Theta_0$ (see next paragraph).

The outer optimization trains the feed-forward network by penalizing novel view rendering errors using the optimized parameters $\Theta^*$:
\begin{equation}\label{eq:outer}
    \mathcal{L}_\text{outer}(\Theta^*) = \mathcal{L}_\text{mse}\left(f(\Theta^*), I_{nv}\right) + \lambda_\text{lpips} \mathcal{L}_\text{LPIPS}\left(f(\Theta^*), I_{nv}\right).
\end{equation}

For each training iteration, the network outputs initial Gaussian parameters $\Theta_0$. These are optimized for a fixed number of steps using \cref{eq:inner} to yield $\Theta^*$. We then compute the outer loss (\cref{eq:outer}) and backpropagate the error to update the feed-forward model parameters $\phi$.

\mypar{Implicit Gradient.}
To update the network weights $\phi$, we must compute the total derivative of the outer loss with respect to $\phi$ using the chain rule:
\begin{equation}
    \frac{\partial \mathcal{L}_\text{outer}}{\partial \phi} = 
    \Big( \frac{\partial \mathcal{L}_\text{outer}}{\partial \Theta^*} \Big)
    \Big( \frac{\partial \Theta^*}{\partial \Theta_0} \Big)
    \Big( \frac{\partial \Theta_0}{\partial \phi} \Big).
\end{equation}
The first term ($\partial \mathcal{L}_\text{outer} / \partial \Theta^*$) is standard rendering backpropagation, and the third term ($\partial \Theta_0 / \partial \phi$) is standard network backpropagation. The central challenge lies in computing the middle term, $\partial \Theta^* / \partial \Theta_0$, which requires differentiating through the inner optimization process. Unrolling the optimization steps is an option but highly memory-intensive. Instead, we compute the exact analytical gradient at the optimum using the Implicit Function Theorem (IFT)~\cite{rajeswaran2019meta, lorraine2020optimizing, bai2019deep}.

At convergence, the inner optimization satisfies the optimality condition, meaning the gradient of the inner loss is zero:
\begin{equation}
    \nabla_\Theta \mathcal{L}_\text{inner}(\Theta^*) = \nabla_\Theta \mathcal{L}_\text{photo}(\Theta^*) + \lambda (\Theta^* - \Theta_0) = \mathbf{0}.
\end{equation}

To find $\frac{\partial \Theta^*}{\partial \Theta_0}$, we differentiate this optimality condition on both sides with respect to $\Theta_0$. Applying the chain rule, we obtain:
\begin{equation}
    \nabla^2_\Theta \mathcal{L}_\text{photo}(\Theta^*) \frac{\partial \Theta^*}{\partial \Theta_0} + \lambda \frac{\partial \Theta^*}{\partial \Theta_0} - \lambda I = \mathbf{0}.
\end{equation}
Let $H_\text{photo} = \nabla^2_\Theta \mathcal{L}_\text{photo}(\Theta^*)$ denote the exact Hessian of the photometric loss. Rearranging the terms yields the closed-form Jacobian of the optimal parameters with respect to the initialization:
\begin{equation}
    \frac{\partial \Theta^*}{\partial \Theta_0} = \lambda (H_\text{photo} + \lambda I)^{-1}.
\end{equation}

Computing the exact Hessian $H_\text{photo}$ involves second-order rendering derivatives, which is computationally prohibitive for 3DGS. Key to our method is to define the photometric loss as a function of the rendering residuals $r(\Theta)$. We can then use the Gauss-Newton approximation to estimate the Hessian as $H_\text{photo} \approx J^\top J$, safely avoiding second-order derivatives. Here, $J = \frac{\partial r(\Theta^*)}{\partial \Theta^*}$ is the Jacobian of the residual with respect to the Gaussian parameters.

Explicitly instantiating and inverting the massive matrix $(J^\top J + \lambda I)$ is intractable. Instead, we formulate this as a linear system by introducing an auxiliary vector $v$:
\begin{equation}
\label{eq:important}
    % (J^\top J + \lambda I) v = \Big(\frac{  \delta\mathcal{L}_{outer} }{ \delta \Theta_0 } \Big)^\top
    (J^\top J + \lambda I) v = \nabla_{\Theta^*} \mathcal{L}_\text{outer}.
\end{equation}
Because the matrix $(J^\top J + \lambda I)$ is symmetric and positive-definite, we can solve for $v$ efficiently using the matrix-free Preconditioned Conjugate Gradient (PCG) algorithm. PCG only requires matrix-vector products.  Once PCG converges to a solution for $v$, the final gradient passed back to the network is simply $\nabla_{\Theta_0} \mathcal{L}_\text{outer} = \lambda v$.

\mypar{Discussion}. It is worth noting the mathematical similarity between our implicit gradient formulation and standard second-order optimization techniques. The left-hand side of our linear system in \cref{eq:important} is structurally identical to the normal equations solved during a Levenberg-Marquardt (LM) optimization step \cite{more2006levenberg}. Therefore, we can leverage the efficient CUDA based LM solver for 3DGS \cite{hollein20253dgs} to speed up our implementation.

In LM, the $(J^\top J + \lambda I)$ is computed for the update steps, where $\lambda$ acts as a damping factor blending Gauss-Newton and gradient descent directions. In our bilevel framework, computing the backward pass of the outer loss through the proximal inner optimization mimics this exact damped step. The proximal weight $\lambda$ from our inner loss elegantly translates into the LM damping factor during the backward pass, ensuring that the gradient flow back to the feed-forward network remains well-conditioned and robust, even when the outer novel-view supervision is noisy or sparse.

Furthermore, analyzing the extreme values of $\lambda$ provides a clear physical intuition for how the gradient behaves during backpropagation:
\begin{itemize}
    \item \textbf{Strong prior (Large $\lambda$):} The optimization is heavily restricted, forcing the final Gaussians $\Theta^*$ to stay near their initialization $\Theta_0$. The gradient passes almost directly through to the network ($\nabla_{\Theta_0} \mathcal{L}_\text{outer} \approx \nabla_{\Theta^*} \mathcal{L}_\text{outer}$), effectively bypassing the inner optimization loop.
    \item \textbf{Weak prior ($\lambda \to 0$):} The Gaussians freely fit the context views and ``forget'' their initialization. Because the final outcome becomes independent of the starting point, the gradient vanishes ($\nabla_{\Theta_0} \mathcal{L}_\text{outer} \to \mathbf{0}$), cutting off the learning signal entirely.
\end{itemize}

% This stark trade-off highlights why a single, global $\lambda$ is insufficient. It perfectly motivates our use of a learnable, per-parameter uncertainty regularizer $\boldsymbol{\Lambda}$, which allows the model to dynamically balance test-time flexibility with a robust gradient flow.

%TODO: Discuss the physical interpretation of the gradient.
% For example if the lambda is big, the optimization cannot go far from the initializatoin and the in the gradient computation part lambda I will domnimate the term making the gradient as original gradients. TODO: in the opposite case

\subsection{Uncertainty-Aware 3DGS Differentiable Optimization}

As highlighted by our physical interpretation, a critical limitation of standard TTO is the use of a rigid, global regularization $\lambda$. It forces a stark compromise: it either overly constrains the test-time adaptation or severs the gradient flow required to train the network. 

Furthermore, a global scalar fails to account for varying confidence levels across the feed-forward predictions; some regions of the scene may be predicted perfectly, while others are highly uncertain. We would like to account for this ``uncertainty'' when regularizing the Gaussian parameters during optimization.

We resolve this by reformulating the inner loss with a per-Gaussian, per-attribute parameter regularizer $\boldsymbol{\Lambda} \in \mathbb{R}^{N_\Theta}$, which are data-dependent meta-parameters predicted alongside $\Theta_0$ by the feed-forward network $g_\phi$:
\begin{equation}
    \mathcal{L}_\text{inner}(\Theta) = \frac{1}{2} \| r(\Theta) \|_2^2 + \frac{1}{2} (\Theta - \Theta_0)^\top \text{diag}(\boldsymbol{\Lambda}) (\Theta - \Theta_0).
\end{equation}
Here, $\boldsymbol{\Lambda}$ replaces the global $\lambda$. If the network is highly uncertain about a specific Gaussian's initial position or color, it predicts a small penalty value in $\boldsymbol{\Lambda}$, allowing that parameter to adapt freely during TTO. Conversely, well-predicted Gaussians are anchored strongly to their initialization $\Theta_0$.

To enable end-to-end training of this uncertainty module, we derive the gradients for the new regularizer by revisiting the Implicit Function Theorem. Because $\boldsymbol{\Lambda}$ is a vector of parameter-specific weights, it acts as a diagonal matrix in our linear system. The modified Hessian approximation becomes $ J^\top J + \text{diag}(\boldsymbol{\Lambda})$. We solve the linear system $(J^\top J + \text{diag}(\boldsymbol{\Lambda}))v = \nabla_{\Theta^*} \mathcal{L}_\text{outer}$ via PCG for $v$, and the modified gradients are:
\begin{equation}
\begin{aligned}
    \nabla_{\Theta_0} \mathcal{L}_\text{outer} &= \boldsymbol{\Lambda} \odot v, \\
    \nabla_{\boldsymbol{\Lambda}} \mathcal{L}_\text{outer} &= - v \odot (\Theta^* - \Theta_0),
\end{aligned}
\end{equation}
where $\odot$ denotes the element-wise product.

These analytical gradients offer a straightforward intuition. Consider the gradient for the uncertainty weights, $\nabla_{\boldsymbol{\Lambda}}$. If the desired update from the outer loss ($v$) points in the same direction as the actual movement during TTO ($\Theta^* - \Theta_0$), their product is positive. The resulting negative gradient reduces $\boldsymbol{\Lambda}$, relaxing the penalty to give that Gaussian more freedom to adapt in future iterations. Through this simple alignment check, the network learns an optimal, per-parameter uncertainty representation directly from novel-view supervision, without requiring explicit ground-truth labels.

% These analytical gradients are highly intuitive. Consider the gradient for the uncertainty weights, $\nabla_{\boldsymbol{\Lambda}}$. If the required optimization update from the novel view ($v$) aligns strongly in the same direction as the actual movement that occurred during TTO ($\Theta^* - \Theta_0$), their product is positive. The negative sign in front then yields a negative gradient for $\boldsymbol{\Lambda}$. This encourages the network to decrease the penalty weight for that parameter in future iterations, progressively loosening the regularization and granting the Gaussian more freedom to move. Consequently, the network learns an optimal, fine-grained uncertainty representation purely from the outer novel-view supervision, entirely without requiring ground-truth uncertainty labels.

\subsection{Implementation}

\mypar{Model}
The proposed methods are agnostic to the design of the feed-forward 3DGS models. In this work, we explore two variations: we build our method on DepthSplat~\cite{xu2025depthsplat} for the feed-forward 3DGS with camera information (\eg, intrinsics and poses), and Depth Anything v3~\cite{lin2025depth} for the setting with multiple images only (\ie, pose-free).

Both methods rely on DPT head~\cite{ranftl2021vision} to predict per-pixel 3D Gaussians and their attributes. Therefore, we add a small MLP to predict Gaussian parameter regularizer $\Lambda$ from the intermediate features in DPT. For both methods, we initialize from the official pretrained weights.

\mypar{Residuals} In practice, rather than a pure L2 loss, we employ a combination of L1 and SSIM losses for the inner TTO objective, following the standard 3DGS optimization, to improve visual fidelity and preserve high-frequency structures. The definition of the residual $r(\Theta)$ and its corresponding Jacobian $J$ are adjusted accordingly via chain rules. 

\mypar{Two-stage training}  Furthermore, to ensure stable convergence and avoid degenerate solutions early in training, we adopt a two-stage training strategy. We first pretrain the feed-forward network $g_\phi$ purely on the outer loss without the differentiable optimization (equivalent to setting $\Theta^* = \Theta_0$). Once the model generates a reasonable initialization, we activate the inner TTO loop and fine-tune the Gaussian DPT head and the uncertainty modules.

To ensure training stability, we apply an auxiliary rendering loss directly to the feed-forward predictions $\Theta_0$, prior to the differentiable optimization layer. Although the optimal initialization for test-time optimization may not perfectly align with the best zero-shot rendering, this proxy loss anchors the network's predictions. Scaled by a down-weighting factor of $\lambda_\text{proxy} = 0.1$, this supervision prevents the model from generating degenerate or unrecoverable Gaussian configurations early in training.

% To make the training more stable, we additionally add the rendering loss  on the output of the model before our differentiable optimization layer scaled by $\lambda_\text{proxy} = 0.1$. Even though the goal of having good initialization is not the same as good rendering output directly, this help the model to maintain reasonable Gaussian initialization, preventing from the training due to unrecoverable initialization.

\mypar{Backward path details.}
We implement our matrix-free PCG solver utilizing the efficient CUDA kernels from 3DGS-LM~\cite{hollein20253dgs}. To ensure stable gradients within the linear system, we normalize the Jacobian matrix $J$ by the total number of rendering residuals. 

In practice, the attributes of a 3D Gaussian (e.g., position, scale, and spherical harmonics) exhibit vastly different scales and gradient magnitudes. 
To properly balance the regularization across these heterogeneous parameter types, we modulate the proximal weight using a Levenberg-Marquardt heuristic. Specifically, the effective proximal term used in our system is formulated as $\lambda \cdot \text{diag}(\boldsymbol{\Lambda} \odot M)$, where $M \in \mathbb{R}^{N_\Theta}$ tracks the moving average of the diagonal of $J^\top J$ during training. This ensures a parameter-aware scaling: attributes that dictate strong changes in the photometric loss, such as the Gaussian means, produce larger values in $M$, thereby receiving a proportionally stronger regularization bound. 

% This technique elegantly harmonizes the scale differences across the Gaussian parameters, stabilizing both the inner test-time optimization and the outer PCG gradient computation.

% We implement our backward PCG solver based on the efficient CUDA kernel from 3DGS-LM~\cite{hollein20253dgs}. In practice,  each Gaussians attributes have different scale and magnitude in the Jacobian matrix. Therefore, we normalize the Jacobian based on the number of the residuals. Also, the weight of our actual proximal term is $\lambda \cdot \text{diag}(\Lambda) \odot \text{diag}(M)$, where $\text{diag}(M)$ is the moving average of the  $\text{diag}(J^\top J)$. For example, moving the Gaussians means would have stronger influence than other attributes, therefore  the diagonal of $J^\top J$ would be larger, making stronger regularization.

\section{Experiments}
\label{sec:exp}

\begin{table}[tb]
  \caption{
    \textbf{Novel view synthesis on RE10K with known poses.}  We report PSNR, SSIM, and LPIPS before and after test-time optimization (TTO) with two input views. While DepthSplat and ours achieve similar zero-shot performance, our method exhibits substantially larger gains after optimization (+0.27 dB vs. +0.03 dB) and attains the highest final scores, demonstrating that our optimization-aware initialization effectively balances prior preservation with adaptation flexibility.
  }
  \label{tab:main_pose_given}
  \setlength{\tabcolsep}{4pt}
  \centering
  \begin{tabular}{@{}lcc@{\hspace{0.5pt}}c@{\hspace{0.5pt}}cccc@{}}
    \toprule
     & \multicolumn{3}{c}{PSNR} & \multicolumn{2}{c}{SSIM} & \multicolumn{2}{c}{LPIPS} \\
     & \scriptsize{Before} & \scriptsize{After} &  & \scriptsize{Before} & \scriptsize{After} & \scriptsize{Before} & \scriptsize{After} \\
    \midrule
    PixelSplat \cite{charatan2024pixelsplat} & 25.856 & 26.144 & \improve{0.288} & 0.848 & 0.854  & 0.150 & 0.145  \\
    MVSplat \cite{chen2024mvsplat} & 26.654 & 26.649 & \worse{0.005} & 0.864 & 0.866 & 0.132 & 0.131 \\
    DepthSplat \cite{xu2025depthsplat} & 27.665 &  27.694 & \improve{0.029} & 0.886 & 0.888 & 0.117 & 0.116 \\
    Ours & \textbf{27.675} & \textbf{27.949} & \improve{0.274} & \textbf{0.887} & \textbf{0.891} & \textbf{0.116} & \textbf{0.112} \\
  \bottomrule
  \end{tabular}
\end{table}

\begin{figure}[tb]
  \centering
  \includegraphics[width=1.0\linewidth]{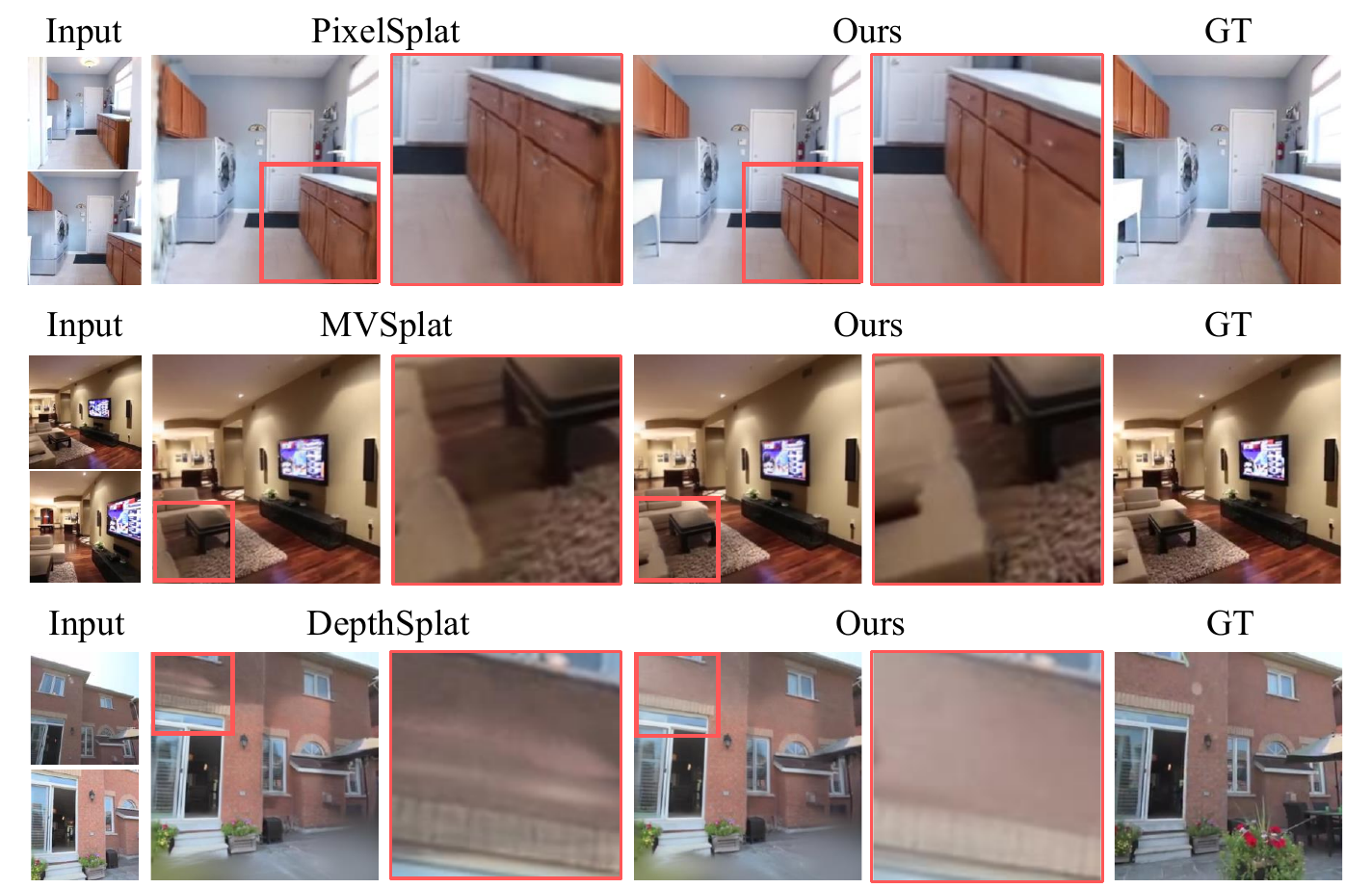}
  \vspace{-4mm}
  \caption{
  \textbf{Qualitative results on RE10K dataset with pose-given setting.} 
  We compare ours with feed-forward 3DGS methods such as PixelSplat~\cite{charatan2024pixelsplat}, MVSplat~\cite{chen2024mvsplat}, and DepthSplat~\cite{xu2025depthsplat} with two input views. As shown in the zoom-in regions, ours generates sharper result, and is more robust to exposure noises in the input views during optimization thanks to the adaptive regularization.
  }
  \label{fig:re10k_compare}
\end{figure}

\mypar{Dataset}
For the pose-given experiments, we train and evaluate the methods on RealEstate10K (RE10K) dataset \cite{zhou2018stereo}. We follow previous methods and use the PixelSplat~\cite{charatan2024pixelsplat} processed train-test split. For the pose-free methods, we fine-tune the Depth Anything v3 using the combination of multiple datasets: RE10K~\cite{zhou2018stereo}, ScanNet~\cite{dai2017scannet}, ScanNet++~\cite{yeshwanth2023scannet++}, DL3DV~\cite{ling2024dl3dv}, Hypersim~\cite{roberts2021hypersim}, and ArkitScenes~\cite{baruch2021arkitscenes}. We evaluate the pose-free methods on RE10K dataset using the same protocol as pose-given experiments, and ScanNet++ on the official validation split.

\mypar{Baselines}
For pose-given setting, we compare our method with PixelSplat~\cite{charatan2024pixelsplat}, MVSplat~\cite{chen2024mvsplat}, and DepthSplat~\cite{xu2025depthsplat} with two input views. For the pose-free experiments, we compare our method with AnySplat~\cite{jiang2025anysplat} and Depth Anything v3 \cite{lin2025depth} with four input views. We use $256\times 256$ resolution for the baselines and our method for all the experiments.

\mypar{Metrics}
We evaluate the standard novel view synthesis metrics: PSNR, SSIM, and LPIPS after running the test-time optimization on the input views. For the pose-given evaluation, we run 50 steps of test-time optimization as we notice that most of the methods have converged after that. For the pose-free evaluation, we firstly align the target poses to the predicted camera poses by estimating the global scale using Umeyama method \cite{umeyama2002least}, and then optimize for the target pose while keeping the 3DGS parameters fixed, following previous works \cite{jiang2025anysplat,ye2025yonosplat,ye2024no} to make sure that the target view poses are aligned with the reconstructed 3D scenes. Then, we run 300 steps of test-time optimization.

\subsection{Novel view synthesis}

\mypar{Pose given.} 
% Experiments results are shown in \cref{tab:main_pose_given} and \cref{fig:re10k_compare}. In general, our method achieves better results than the baselines after test-time optimization. The reason is that our model predicts a initialization that might be worse in terms of rendering quality in the output, but are easier to optimize. Furthermore, the proximal term with adaptive regularization make sure that the model are less affected by overfitting under sparse settings. For example, in last row \cref{fig:re10k_compare}, the baseline method suffers from the color changes in the input, while our method could counter that dramatic color changes during test-time optimization by learning from data, as the method are trained to provide the best novel view rendering after optimization.
Quantitative and qualitative results are presented in \cref{tab:main_pose_given} and \cref{fig:re10k_compare}, respectively. Our method consistently outperforms baseline approaches following test-time optimization (TTO). This performance gain is fundamentally driven by our bilevel formulation: rather than strictly maximizing zero-shot rendering fidelity, the network learns to predict optimization-aware initializations that are more favorable for TTO. Furthermore, our adaptive proximal regularizer ensures the optimization process is highly robust against overfitting in under-constrained, sparse-view settings. For instance, as shown in the bottom row of \cref{fig:re10k_compare}, baseline methods degrade severely when exposed to inconsistent color shifts in the context views. Conversely, our approach mitigates these dramatic illumination changes during TTO. By training end-to-end to maximize post-optimization novel view quality, our model learns data-dependent uncertainty bounds that effectively discount such unreliable observations.

\begin{table}[tb]
  \caption{
    \textbf{Pose-free feed-forward 3DGS on RE10K.} Comparison of feed-forward methods using four input views without camera information. We report PSNR, SSIM, and LPIPS before and after test-time optimization (TTO). Notably, AnySplat suffers from catastrophic overfitting (PSNR degrades by 0.2 dB) due to poor initialization, while our method achieves the largest performance gains (+0.76 dB vs. +0.60 dB for DA3) and attains the best final scores across all metrics.
  }
  \label{tab:re10k_pose_free}
  \setlength{\tabcolsep}{4pt}
  \centering
  \begin{tabular}{@{}lcc@{\hspace{0.5pt}}c@{\hspace{0.5pt}}cccc@{}}
    \toprule
     & \multicolumn{3}{c}{PSNR} & \multicolumn{2}{c}{SSIM} & \multicolumn{2}{c}{LPIPS} \\
     & \scriptsize{Before} & \scriptsize{After} &  & \scriptsize{Before} & \scriptsize{After} & \scriptsize{Before} & \scriptsize{After} \\
    \midrule
    % NopoSplat \cite{ye2024no} & \\
    % Spatt3R \cite{smart2024splatt3r} & \\
    AnySplat \cite{jiang2025anysplat} & 17.762 & 17.546 & \worse{0.216} & 0.603 & 0.585 & 0.325 & 0.328  \\
    DA3 \cite{lin2025depth} & 21.981 & 22.582 & \improve{0.601} & 0.741 & 0.764 & 0.223 & 0.205 \\
    Ours & \textbf{22.043} & \textbf{22.806} & \improve{0.763} & \textbf{0.742} & \textbf{0.767} & \textbf{0.221} & \textbf{0.202}  \\
  \bottomrule
  \end{tabular}
\end{table}

\begin{table}[tb]
  \caption{
    \textbf{Pose-free evaluation on ScanNet++ (DSLR \& iPhone) with four input views.} Results on the official validation split using four unposed input views. We compare against AnySplat and Depth Anything v3 (DA3), reporting metrics before and after TTO. Our method consistently outperforms baselines after optimization across both capture modalities, achieving particularly strong gains on the challenging DSLR split (+0.35 dB PSNR against DA3).
  }
  \label{tab:scannet_pose_free}
  \centering
  \setlength{\tabcolsep}{1.5pt}
  \scriptsize
  \begin{tabular}{@{}lccccccccccccc@{}}
    \toprule
    &  \multicolumn{6}{c}{ScanNet++ (DSLR)} &  \multicolumn{6}{c}{ScanNet++ (iPhone)} \\
    \cmidrule{2-7} \cmidrule{9-14}
     & \multicolumn{2}{c}{PSNR} & \multicolumn{2}{c}{SSIM} & \multicolumn{2}{c}{LPIPS} && \multicolumn{2}{c}{PSNR} & \multicolumn{2}{c}{SSIM} & \multicolumn{2}{c}{LPIPS} \\
     & Before & After & Before & After & Before & After & & Before & After & Before & After & Before & After  \\
    \midrule
    AnySplat \cite{jiang2025anysplat} & 18.84 & 18.66 & 0.675 & 0.654 & 0.307 & 0.312 && 18.21 & 18.08 & 0.671 &  0.655 & 0.304 & 0.299 \\
    DA3 \cite{lin2025depth} & 23.54 & 24.29 & 0.820 & 0.844 & 0.193 & 0.174 && \textbf{23.23} & 24.43 & \textbf{0.840} & 0.878 & 0.171 & 0.152 \\
    Ours & \textbf{23.78} & \textbf{24.64} & \textbf{0.824} & \textbf{0.850} & \textbf{0.182} & \textbf{0.162} && 23.15 & \textbf{24.57} & \textbf{0.840} & \textbf{0.881} & \textbf{0.167} & \textbf{0.145} \\
  \bottomrule
  \end{tabular}
  
\end{table}

\begin{figure}[tb]
  \centering
  \includegraphics[width=1.0\linewidth]{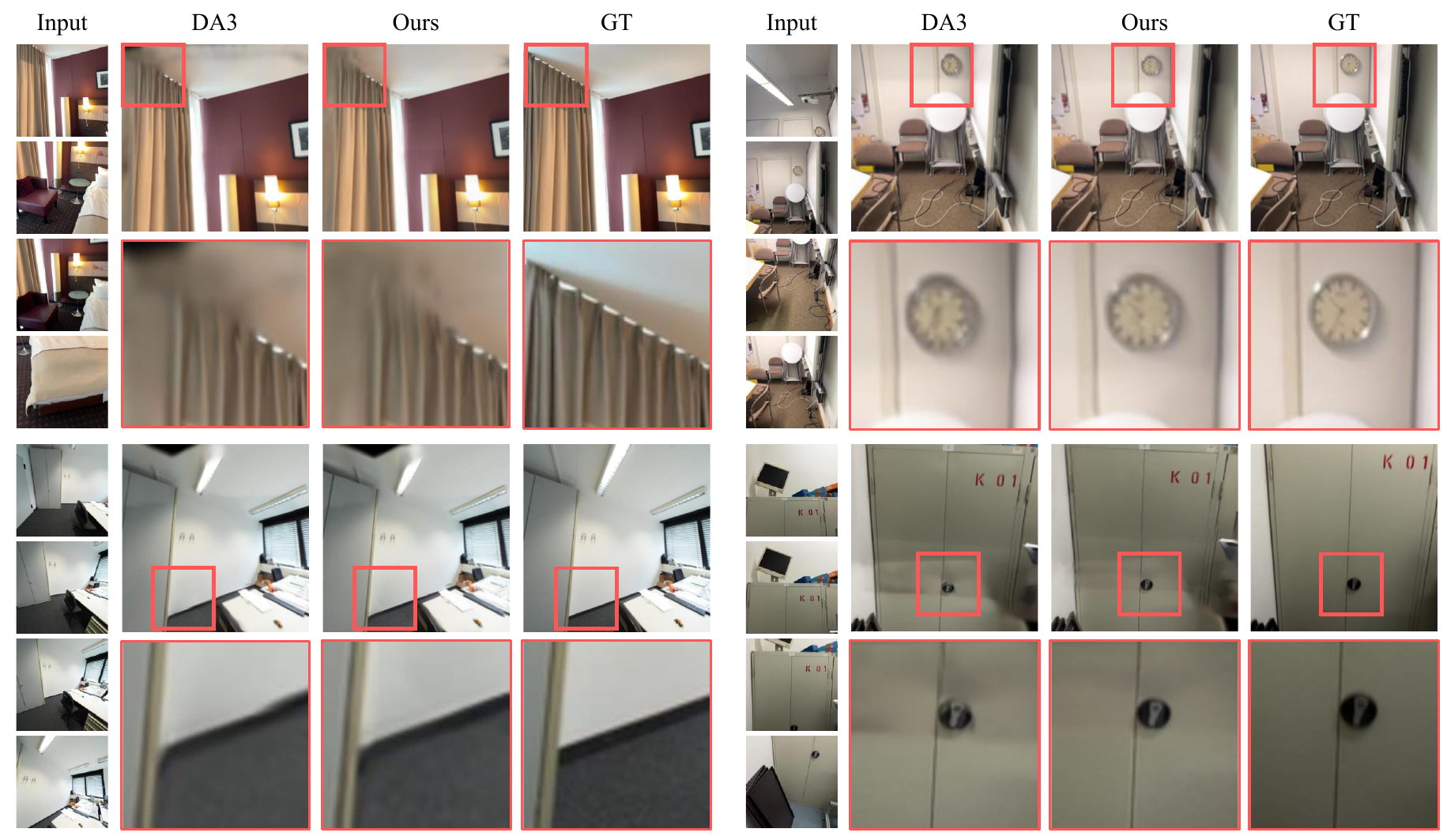}
  \vspace{-4mm}
  \caption{\textbf{Qualitative results on ScanNet++.} We compare our pose-free methods against Depth Anything v3 (DA3) \cite{lin2025depth} after optimization on ScanNet++ with four input views. Our method is able to reduce the artifacts during optimization and provides sharper renderings.
  }
  \label{fig:scannetpp_compare}
\end{figure}

\mypar{Pose free.} 
% The experiments with pose-free baselines are presented in \cref{tab:re10k_pose_free} (RE10K) and \cref{tab:scannet_pose_free} (ScanNet++). Our method successfully boost the performance of Depth Anything v3 after test-time optimization. 
For pose-free reconstruction, we build upon Depth Anything v3 (DA3) as our base feed-forward model. As shown in \cref{tab:re10k_pose_free} (RE10K) and \cref{tab:scannet_pose_free} (ScanNet++), our uncertainty-aware optimization outperforms standard TTO applied to vanilla DA3, achieving larger performance gains after optimization (+0.76 dB vs. +0.60 dB on RE10K) while producing higher-quality initializations.
As shown in \cref{fig:scannetpp_compare}, \OURS{} can generate sharper details with fewer artifacts caused by overfitting to the input views.
% \jozef{I'd describe the results here a bit as well, e.g. For pose-free reconstruction, we build upon Depth Anything v3 (DA3) as our base feed-forward model. As shown in \cref{tab:re10k_pose_free} (RE10K) and \cref{tab:scannet_pose_free} (ScanNet++), our uncertainty-aware optimization outperforms standard TTO applied to vanilla DA3, achieving larger performance gains after optimization (+0.76 dB vs. +0.60 dB on RE10K) while producing higher-quality initializations.} \yc{merged. thanks}

\subsection{Ablations}

\begin{table}[tb]
  \caption{\textbf{Ablation study on \OURS{}.} 
  We compare our full method and other variations on RE10K, pose-given, two-view settings. Without proximal term, the method is still benefiting from learning a better initialization but becomes worse due to unconstrained optimization. On the other hand, without the learnable parameter regularization, the global bound makes the optimization too conservative.
  }
  \label{tab:ablation}
  \setlength{\tabcolsep}{2pt}
  \centering
  \begin{tabular}{@{}llcc@{\hspace{0.5pt}}c@{\hspace{0.5pt}}cccc@{}}
    \toprule
     && \multicolumn{3}{c}{PSNR} & \multicolumn{2}{c}{SSIM} & \multicolumn{2}{c}{LPIPS} \\
    & & Before & After & & Before & After & Before & After \\
    \midrule
    (a) & DepthSplat  \cite{xu2025depthsplat} & 27.665 &  27.694 & \improve{0.029} & 0.886 & 0.888 & 0.117 & 0.116 \\
    % (b) & Ours w/o $\mathcal{L}_\text{proximal}$ & 27.598 & 27.759 & \improve{0.161} & 0.885 & 0.889 & 0.117 & 0.116 \\
    (b) & Ours w/o $\mathcal{L}_\text{proximal}$ &\textbf{27.675} & 27.828 & \improve{0.153} & 0.887 & 0.890 & \textbf{0.116}  & 0.115 \\
    % (c) & Ours w/o $\Lambda$ & 27.630 & 27.746 & \improve{0.116} & \textbf{0.887} & 0.889 & \textbf{0.116} & 0.115 \\
    (c) & Ours w/o $\Lambda$ & 27.630 & 27.746 & \improve{0.116} & \textbf{0.888} & 0.889 & \textbf{0.116} & 0.115 \\
    % (d)& Ours & 27.579 & \textbf{27.880} & \improve{0.301} & 0.885 & \textbf{0.890} & 0.117 & \textbf{0.113} \\
    (d) & Ours & \textbf{27.675} & \textbf{27.949} & \improve{0.274} & 0.887 & \textbf{0.891} & \textbf{0.116} & \textbf{0.112} \\
    
  \bottomrule
  \end{tabular}
\end{table}

\paragraph{Is the proximal term effective?}
We drop the proximal term during test-time optimization after training. The result is shown in \cref{tab:ablation} (b). Compared to vanilla DepthSplat it still improves due to better initialization. However, it is worse than the full method (d) since the optimization is less constrained and still suffers from overfitting to the input views.

\paragraph{Are the learnable regularization weights effective?} The parameters are bounded by a single lambda value uniformly without the learnable regularization weights. Therefore, the optimization becomes more conservative. As shown in \cref{tab:ablation} (c), the initializations are forced to behave like a standard feed-forward, and the improvements from the optimization are limited.

\begin{figure}[tb]
  \centering
  \includegraphics[width=1.0\linewidth]{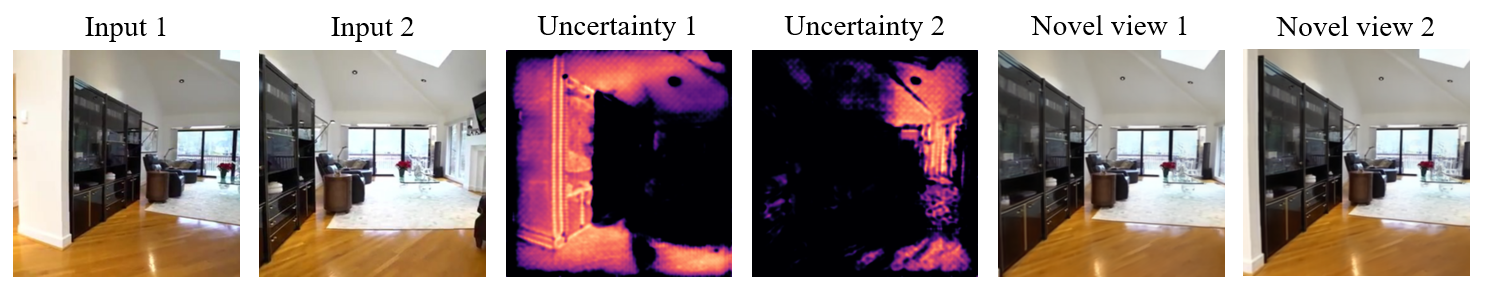}
  \vspace{-4mm}
  \caption{\textbf{Visualization of learned uncertainty weights.} We visualize the regularization weights applied to the mean of the per-pixel Gaussians ($\Lambda_\text{mean}$).
  Notably, the network intrinsically learns to predict higher anchoring weights for regions lacking multi-view constraints (\eg, the left third of Input 1, which is not visible in Input 2) without explicit supervision. This helps to prevent overfitting during the test-time optimization process.
  }
  \label{fig:uncertainty}
\end{figure}

\subsection{Limitation} Similar to differentiable bundle-adjustment in VGG-SfM \cite{wang2024vggsfm}, a notable limitation in our differentiable formulation is increased training time. This increases the training time for each step from $4\times$ to $5\times$ compared to the original feed-forward training. Note that our training implementation remains unoptimized for speed (where the forward pass constitutes the majority of the time, as it runs the gradient descent for 50-100 steps or 65\% of the additional time), and there is no resulting increase in runtime at test time. 
%The main bottleneck of our differentiable 3DGS optimization is the forward pass which runs the gradient descent (\eg, Adam) for 50-100 steps  (65\% of the additionally time) and the backward pass takes roughly 35\% of the additional time time. However, it does not increase any overhead during test-time.
\section{Conclusion}
\label{sec:conclusion}

% In this work, we presented \OURS{}, a novel method that bridges the gap between the speed of feed-forward 3D Gaussian Splatting and the high-fidelity reconstruction of per-scene test-time optimization. The core of our method is the efficient differentiable 3DGS optimization layer based on the Implicit Function Theorem and a matrix-free PCG solver. This enables us to train feed-forward 3DGS models combined with test-time optimization, making the network learn to predict ``better initialization'' downstream optimization. Furthermore, we show that by modeling uncertainty with the learnable regularization weights, our model could determine the strength of anchoring for each parameter that is best for the optimization adaptively. Although that training process would be computationally slower, we believe that this differentiable 3DGS optimization could inspire future works to work on learning to refine 3DGS on-the-fly (\eg, 3DGS SLAM) or extend the uncertainty to other data like 4D scenes or transient objects.

In this work, we presented \OURS{}, a novel framework that bridges the speed of feed-forward 3D Gaussian Splatting and the high-fidelity reconstruction of per-scene test-time optimization (TTO). At its core is an efficient, differentiable 3DGS optimization layer powered by the Implicit Function Theorem and a matrix-free PCG solver. This formulation enables end-to-end training through the TTO process, driving the network to predict an optimal initialization specifically tailored for downstream refinement. Furthermore, we showed that by modeling uncertainty through learnable regularization weights, our framework adaptively determines the anchoring strength for each Gaussian parameter, effectively preventing overfitting to sparse context views. Although this bilevel training process is computationally heavier than standard feed-forward approaches, our differentiable formulation opens exciting avenues for future research. We believe this work can inspire new methods for learning to refine 3DGS on-the-fly (\eg, in 3DGS SLAM) or extending uncertainty-aware optimization to dynamic 4D scenes and transient objects.

\section*{Acknowledgements}

This work was supported by the ERC Starting Grant SpatialSem (101076253), the ERC Consolidator Grant Gen3D (101171131), and by the Computing Systems Lab, part of the Huawei Technologies Switzerland AG.
The authors gratefully acknowledge the Gauss Centre for Supercomputing e.V. (www.gauss-centre.eu) for funding this project by providing computing time through the John von Neumann Institute for Computing (NIC) on the GCS Supercomputer JUWELS \cite{JUWELS} at Jülich Supercomputing Centre (JSC).
We also acknowledge the EuroHPC Joint Undertaking for providing access to the EuroHPC supercomputer MareNostrum 5, hosted by the Barcelona Supercomputing Center (BSC), Spain. 
Finally, we would like to thank Lukas H{\"o}llein for his valuable support with the 3DGS-LM code.

% \section{Introduction}
% \label{sec:intro}

% \clearpage

\bibliographystyle{splncs04}
\bibliography{main}

\clearpage
\appendix

\section{Additional Qualitative Results}

\begin{figure}[htb]
  \centering
  \includegraphics[width=1.0\linewidth]{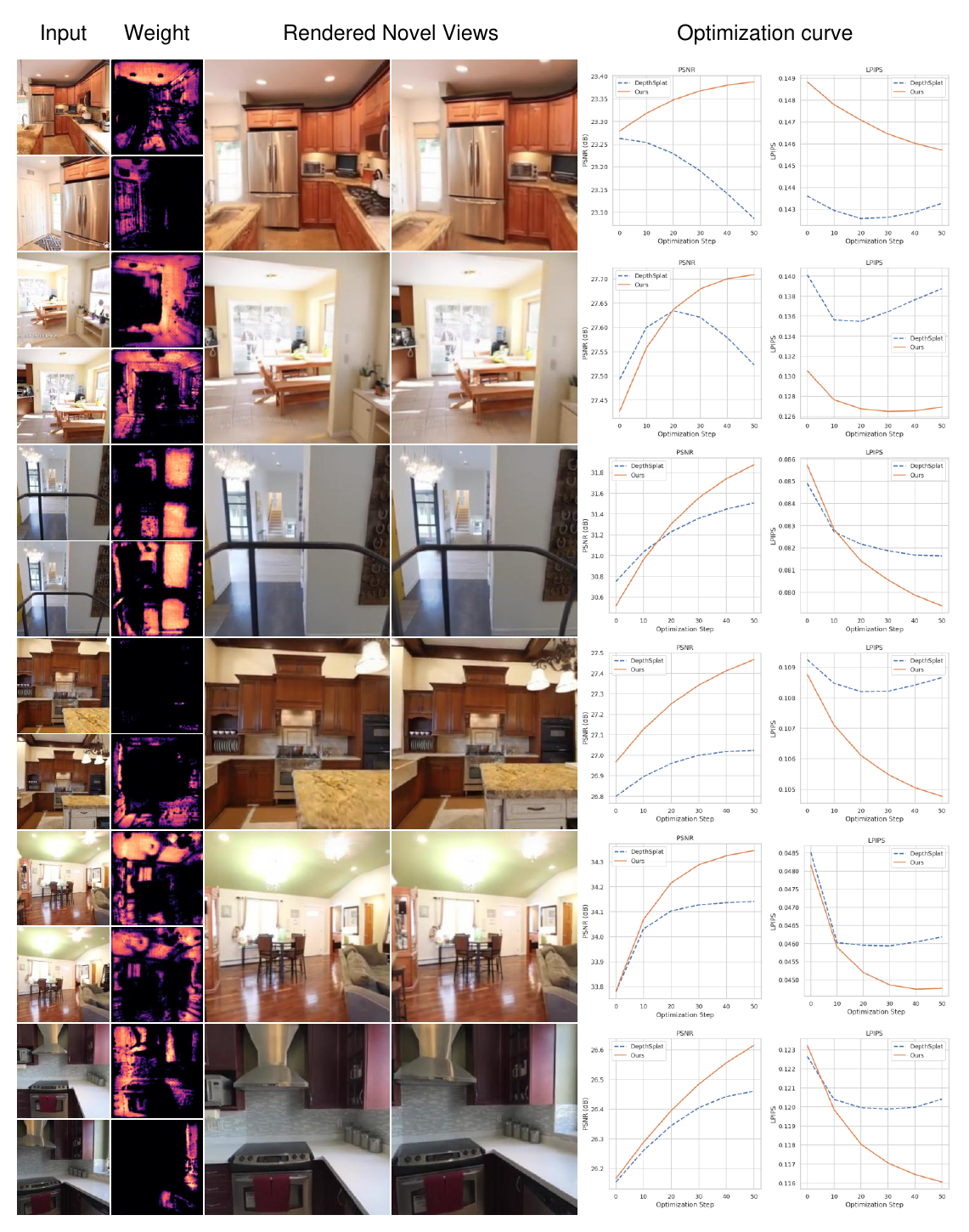}
  \vspace{-4mm}
  \caption{
  \textbf{More qualitative results on RE10K.}
    Our predicted uncertainty weights implicitly learn to penalize changes (\ie, higher values) in non-overlapping regions during post-optimization. Crucially, we show that \OURS{} learns a more effective initialization: while the initial predictions may yield worse zero-shot renderings, they result in higher-quality outputs following optimization.
    % We show that our uncertainty weights implicitly learns to reduce the changes (\ie, higher values) in the regions without overlapping in the input views during optimization. Together with the differentiable optimization, we should that \OURS{} can produce better rendering after optimization even with worse initialization.
  }
  \label{fig:more_results_suppl}
\end{figure}

We provide additional qualitative results for the two-view, pose-given setting on the RE10K dataset in \cref{fig:more_results_suppl}. Visualizations of the predicted uncertainty weights demonstrate that \OURS{} intrinsically learns to penalize parameter updates in regions lacking multi-view overlap. This adaptive regularization effectively anchors uncertain Gaussians, making the test-time optimization significantly more robust against overfitting to the input views. 

Furthermore, the accompanying optimization curves illustrate our core idea. Rather than strictly maximizing zero-shot rendering quality, \OURS{} predicts an optimization-aware initialization. By starting in a more favorable region of the loss landscape, our method consistently converges to a preferable local minimum, ultimately achieving higher-fidelity post-optimization renderings compared to the baseline.

\section{Detailed Derivation of the Implicit Gradient}
\label{supp:implicit_gradient}

In this section, we provide the  derivation details of the implicit gradient used to backpropagate the outer loss through the inner test-time optimization (TTO) process. 

\subsection{Optimality Condition and Implicit Function Theorem}
Recall that the inner optimization objective is defined as:
\begin{equation}
    \mathcal{L}_\text{inner}(\Theta) = \mathcal{L}_\text{photo}(\Theta) + \frac{\lambda}{2} \| \Theta - \Theta_0 \|_2^2,
\end{equation}
where $\mathcal{L}_\text{photo}(\Theta) = \frac{1}{2} \| r(\Theta) \|_2^2$ is the photometric loss based on the rendering residual $r(\Theta)$, $\Theta_0$ is the feed-forward initialization, and $\lambda$ is the proximal regularization weight.

We assume the inner optimization converges to a local minimum $\Theta^*$. At this optimum, the derivative of the inner loss with respect to the Gaussian parameters $\Theta$ must be exactly zero. This gives us the optimality condition:
\begin{equation}
\label{eq:supp_optimality}
    \nabla_\Theta \mathcal{L}_\text{inner}(\Theta^*) = \nabla_\Theta \mathcal{L}_\text{photo}(\Theta^*) + \lambda (\Theta^* - \Theta_0) = \mathbf{0}.
\end{equation}

To compute how a change in the initialization $\Theta_0$ affects the optimal parameters $\Theta^*$, we differentiate both sides of \cref{eq:supp_optimality} with respect to $\Theta_0$. Applying the multivariate chain rule yields:
\begin{equation}
    \nabla^2_\Theta \mathcal{L}_\text{photo}(\Theta^*) \frac{\partial \Theta^*}{\partial \Theta_0} + \lambda \frac{\partial \Theta^*}{\partial \Theta_0} - \lambda I = \mathbf{0},
\end{equation}
where $I$ is the identity matrix. Let $H_\text{photo} = \nabla^2_\Theta \mathcal{L}_\text{photo}(\Theta^*)$ denote the exact Hessian of the photometric loss. We can factor out the Jacobian $\frac{\partial \Theta^*}{\partial \Theta_0}$:
\begin{equation}
    (H_\text{photo} + \lambda I) \frac{\partial \Theta^*}{\partial \Theta_0} = \lambda I.
\end{equation}

Assuming the regularized Hessian $(H_\text{photo} + \lambda I)$ is invertible (which is guaranteed for a sufficiently large proximal weight $\lambda$ as it strictly convexifies the local landscape), we obtain the closed-form Jacobian of the optimal parameters with respect to the initialization:
\begin{equation}
\label{eq:supp_jacobian_exact}
    \frac{\partial \Theta^*}{\partial \Theta_0} = \lambda (H_\text{photo} + \lambda I)^{-1}.
\end{equation}

\subsection{Gauss-Newton Approximation}
Computing the exact Hessian $H_\text{photo}$ requires second-order derivatives of the rendering process, which is computationally intractable for 3D Gaussian Splatting. Because the photometric loss is a sum-of-squares residual objective ($\frac{1}{2} \| r(\Theta) \|_2^2$), we can employ the Gauss-Newton approximation.

By definition, the exact Hessian is $H_\text{photo} = J^\top J + \sum_i r_i \nabla^2 r_i$, where $J = \frac{\partial r(\Theta^*)}{\partial \Theta^*}$ is the Jacobian of the residual. Assuming the residuals $r_i$ are small near the optimum, we can drop the second-order term, yielding:
\begin{equation}
    H_\text{photo} \approx J^\top J.
\end{equation}
Substituting this back into \cref{eq:supp_jacobian_exact}, we get our workable Jacobian formulation:
\begin{equation}
\label{eq:supp_jacobian_gn}
    \frac{\partial \Theta^*}{\partial \Theta_0} \approx \lambda (J^\top J + \lambda I)^{-1}.
\end{equation}

\subsection{Vector-Jacobian Product and PCG Objective}
To update the feed-forward network weights, we need the gradient of the outer loss $\mathcal{L}_\text{outer}$ with respect to $\Theta_0$. Using the chain rule, this is the Vector-Jacobian Product (VJP) between the outer loss gradient and the implicit Jacobian:
\begin{equation}
\label{eq:supp_vjp}
    \nabla_{\Theta_0} \mathcal{L}_\text{outer} = \frac{\partial \Theta^*}{\partial \Theta_0}^\top \nabla_{\Theta^*} \mathcal{L}_\text{outer} = \lambda (J^\top J + \lambda I)^{-1} \nabla_{\Theta^*} \mathcal{L}_\text{outer}.
\end{equation}
Note that $(J^\top J + \lambda I)$ is symmetric, so its transpose is itself.

Explicitly instantiating and inverting the matrix $(J^\top J + \lambda I)$ is impossible since it scales quadratically with the number of Gaussian parameters ($N \times N$, where $N$ is often in the millions). Instead, we treat the term $(J^\top J + \lambda I)^{-1} \nabla_{\Theta^*} \mathcal{L}_\text{outer}$ as an unknown vector $v$.

We define $v$ such that:
\begin{equation}
    v = (J^\top J + \lambda I)^{-1} \nabla_{\Theta^*} \mathcal{L}_\text{outer}.
\end{equation}
Multiplying both sides by the regularized Hessian transforms the matrix inversion problem into a large-scale system of linear equations:
\begin{equation}
\label{eq:supp_pcg}
    (J^\top J + \lambda I)v = \nabla_{\Theta^*} \mathcal{L}_\text{outer}.
\end{equation}

Once PCG converges to a solution for $v$, substituting it back into \cref{eq:supp_vjp} yields the final, exact gradient to be passed back to the network:
\begin{equation}
    \nabla_{\Theta_0} \mathcal{L}_\text{outer} = \lambda v.
\end{equation}

\subsection{Gradients for the Uncertainty Weights}

When transitioning from a global scalar $\lambda$ to the per-parameter uncertainty regularizer $\boldsymbol{\Lambda} \in \mathbb{R}^{N_\Theta}$, the inner loss becomes:
\begin{equation}
    \mathcal{L}_\text{inner}(\Theta) = \mathcal{L}_\text{photo}(\Theta) + \frac{1}{2} (\Theta - \Theta_0)^\top \text{diag}(\boldsymbol{\Lambda}) (\Theta - \Theta_0).
\end{equation}

At convergence, the new optimality condition is:
\begin{equation}
\label{eq:supp_optimality_lambda}
    \nabla_\Theta \mathcal{L}_\text{inner}(\Theta^*) = \nabla_\Theta \mathcal{L}_\text{photo}(\Theta^*) + \text{diag}(\boldsymbol{\Lambda}) (\Theta^* - \Theta_0) = \mathbf{0}.
\end{equation}

To find the implicit gradient of the optimal parameters with respect to the uncertainty weights, we differentiate this optimality condition with respect to $\boldsymbol{\Lambda}$. Applying the product rule yields:
\begin{equation}
    \nabla^2_\Theta \mathcal{L}_\text{photo}(\Theta^*) \frac{\partial \Theta^*}{\partial \boldsymbol{\Lambda}} + \text{diag}(\boldsymbol{\Lambda}) \frac{\partial \Theta^*}{\partial \boldsymbol{\Lambda}} + \text{diag}(\Theta^* - \Theta_0) = \mathbf{0}.
\end{equation}
Substituting the Gauss-Newton approximation $H_\text{photo} \approx J^\top J$ and rearranging the terms allows us to solve for the Jacobian $\frac{\partial \Theta^*}{\partial \boldsymbol{\Lambda}}$:
\begin{equation}
\label{eq:supp_jacobian_lambda}
    (J^\top J + \text{diag}(\boldsymbol{\Lambda})) \frac{\partial \Theta^*}{\partial \boldsymbol{\Lambda}} = - \text{diag}(\Theta^* - \Theta_0)
\end{equation}
\begin{equation}
    \frac{\partial \Theta^*}{\partial \boldsymbol{\Lambda}} = - (J^\top J + \text{diag}(\boldsymbol{\Lambda}))^{-1} \text{diag}(\Theta^* - \Theta_0).
\end{equation}

Next, we apply the chain rule to compute the gradient of the outer loss with respect to $\boldsymbol{\Lambda}$. This requires multiplying the transpose of our Jacobian by the outer loss gradient:
\begin{equation}
    \nabla_{\boldsymbol{\Lambda}} \mathcal{L}_\text{outer} = \left( \frac{\partial \Theta^*}{\partial \boldsymbol{\Lambda}} \right)^\top \nabla_{\Theta^*} \mathcal{L}_\text{outer}.
\end{equation}

Substituting \cref{eq:supp_jacobian_lambda} into this equation gives:
\begin{equation}
    \nabla_{\boldsymbol{\Lambda}} \mathcal{L}_\text{outer} = - \left( (J^\top J + \text{diag}(\boldsymbol{\Lambda}))^{-1} \text{diag}(\Theta^* - \Theta_0) \right)^\top \nabla_{\Theta^*} \mathcal{L}_\text{outer}.
\end{equation}

Because the approximated Hessian $(J^\top J + \text{diag}(\boldsymbol{\Lambda}))$ is symmetric, its inverse is also symmetric. Likewise, the diagonal matrix $\text{diag}(\Theta^* - \Theta_0)$ is symmetric. Thus, the transpose operation simply reverses the order of the matrices without altering their contents:
\begin{equation}
    \nabla_{\boldsymbol{\Lambda}} \mathcal{L}_\text{outer} = - \text{diag}(\Theta^* - \Theta_0) \underbrace{(J^\top J + \text{diag}(\boldsymbol{\Lambda}))^{-1} \nabla_{\Theta^*} \mathcal{L}_\text{outer}}_{v}.
\end{equation}

Notice that the rightmost term is exactly the auxiliary vector $v$ we already solved for via the matrix-free PCG solver (adapted from \cref{eq:supp_pcg} to use the diagonal $\boldsymbol{\Lambda}$ matrix):
\begin{equation}
    v = (J^\top J + \text{diag}(\boldsymbol{\Lambda}))^{-1} \nabla_{\Theta^*} \mathcal{L}_\text{outer}.
\end{equation}

Substituting $v$ back into the equation simplifies the expression significantly:
\begin{equation}
    \nabla_{\boldsymbol{\Lambda}} \mathcal{L}_\text{outer} = - \text{diag}(\Theta^* - \Theta_0) v.
\end{equation}

Finally, multiplying a diagonal matrix by a column vector is mathematically equivalent to computing the element-wise  product of the two vectors. This yields our final, highly efficient analytical gradient:
\begin{equation}
    \nabla_{\boldsymbol{\Lambda}} \mathcal{L}_\text{outer} = - v \odot (\Theta^* - \Theta_0).
\end{equation}

By similar logic, the gradient for the initialization $\Theta_0$ under the per-parameter formulation cleanly updates to $\nabla_{\Theta_0} \mathcal{L}_\text{outer} = \boldsymbol{\Lambda} \odot v$. Because both gradients reuse the exact same PCG solution $v$, computing the uncertainty gradients incurs virtually zero additional computational overhead.

% \begin{figure}[tb]
%   \centering
%   % \includegraphics[height=6.5cm]{eijkel2}
%     \fbox{\rule{0pt}{0.5in} \rule{.9\linewidth}{0pt}}
%   \caption{Visualize our method (RGB, depth, uncertainty weights)}
%   \label{fig:visualize}
% \end{figure}

% \begin{table}[tb]
%   \caption{Evaluation of optimization based for sparse views}
%   \label{tab:main_sparse}
%   \centering
%   \setlength{\tabcolsep}{0.5em} % for the horizontal padding
%   \begin{tabular}{@{}llll@{}}
%     \toprule
%      & PSNR & SSIM & LPIPS  \\
%     \midrule
%     FSGS (Depth reg sparse 3DGS) & \\
%     InstantSplat & \\
%     Ours & \\
%   \bottomrule
%   \end{tabular}
% \end{table}

% \begin{figure}[tb]
%   \centering
%   % \includegraphics[height=6.5cm]{eijkel2}
%     \fbox{\rule{0pt}{0.5in} \rule{.9\linewidth}{0pt}}
%   \caption{Showcase the optimization curve between baselines and ours. Show case the baseline overfits while ours be able to fight against that. Optional: try completely 0 overlap pairs}
%   \label{fig:convergence}
% \end{figure}

% \begin{table}[tb]
%   \caption{Ablation on the methods}
%   \label{tab:ablation}
%   \centering
%   \begin{tabular}{@{}lllllll@{}}
%     \toprule
%      & \multicolumn{3}{c}{num views = 2} & \multicolumn{3}{c}{num views = 4} \\
%       & PSNR & SSIM & LPIPS & PSNR & SSIM & LPIPS \\
%     w/o diff opt & \\
%     w/o diff opt + only optimize the color \\
%     w/o diff opt + manual params reg & \\
%     First order & \\
%     w/ diff opt &  \\
%     w/ diff opt + reg & \\
%     w/ diff opt + ? & \\
%   \bottomrule
%   \end{tabular}
% \end{table}

\end{document}